\documentclass[EPiC]{easychair}

\usepackage{doc}

\usepackage{amsmath}
\usepackage{amssymb}
\usepackage{amsfonts}
\usepackage{dsfont}
\usepackage{microtype}
\usepackage{graphicx}
\usepackage{subfigure}
\usepackage{booktabs}
\usepackage{hyperref}
\usepackage{graphicx}
\usepackage[numbers,sort&compress]{natbib}
\usepackage[utf8]{inputenc}
\usepackage[printonlyused,nolist]{acronym}

\newcommand{\EE}[2][]{\mathbb{E}_{#1}\left[#2\right]}

\newcommand{\IR}{\mathds{R}}

\newcommand{\set}[1]{\left\{#1\right\}}

\title{Multi-task Learning and Catastrophic Forgetting in\\
Continual Reinforcement Learning}

\titlerunning{Multitask Learning and Forgetting in Reinforcement Learning}

\author{
    João Ribeiro
\and
    Francisco S.\ Melo
\and
   João Dias
}

\institute{
   INESC-ID/Instituto Superior T\'{e}cnico\\
   University of Lisbon\\
   Lisbon, Portugal
 }

\authorrunning{Ribeiro, Melo, and Dias}

\titlerunning{Multi-task learning and catastrophic forgetting in \\
continual reinforcement learning}

\begin{document}

\maketitle

\begin{abstract}
In this paper we investigate two hypothesis regarding the use of deep reinforcement learning in multiple tasks. The first hypothesis is driven by the question of whether a deep reinforcement learning algorithm, trained on two similar tasks, is able to outperform two single-task, individually trained algorithms, by more efficiently learning a new, similar task, that none of the three algorithms has encountered before. The second hypothesis is driven by the question of whether the same multi-task deep RL algorithm, trained on two similar tasks and augmented with elastic weight consolidation (EWC), is able to retain similar performance on the new task, as a similar algorithm without EWC, whilst being able to overcome catastrophic forgetting in the two previous tasks. We show that a multi-task Asynchronous Advantage Actor-Critic (GA3C) algorithm, trained on Space Invaders and Demon Attack, is in fact able to outperform two single-tasks GA3C versions, trained individually for each single-task, when evaluated on a new, third task---namely, Phoenix. We also show that, when training two trained multi-task GA3C algorithms on the third task, if one is augmented with EWC, it is not only able to achieve similar performance on the new task, but also capable of overcoming a substantial amount of catastrophic forgetting on the two previous tasks.%
\footnote{Our source code and obtained results are publicly available at \texttt{https://github.com/jmribeiro/UGP}.}
\end{abstract}




\section{Introduction}
\label{sect:introduction}

Recent years have witnessed a significant number of reinforcement learning successes \cite{hafner2018learning,barth2018distributed,schulman2017proximal,babaeizadeh2016reinforcement,silver2016mastering,tesauro1995temporal}, many of which can be credited to the use of deep neural networks \cite{lecun2015deep,lecun2010convolutional} in the context of reinforcement learning (RL) problems, in what has become known as {\em deep reinforcement learning}. Until the advent of deep RL, the success of RL agents in complex domains depended largely on a careful design of features by the agent designers \cite{boyan94nips,sutton95nips,parr08icml,nguyen13icml}. 
In contrast, deep RL agents directly can learn from large unprocessed inputs, overcoming the need for the agent designer to specify which features are relevant for the task at hand. For example, deep $Q$-networks (DQN) \cite{mnih2013playing,mnih2015human} successfully learned a number of different tasks in the Atari2600 platform \cite{bellemare2013arcade} from only raw pixel inputs and reward information, without manually designed features.

However, although the same architecture can be used to learn different tasks---as was the case with the aforementioned DQNs---each different task was learned with a different instance of DQN. When faced with a new task after learning a previous one, the DQN agent would have to learn the second task from scratch, even if both tasks were very similar. Even though there have been some agents that were able to successfully learn multiple tasks \cite{caruana1998multitask,birk2017hybrida3c}, the ability to transfer knowledge between tasks remains an important challenge in deep RL. 

A second related limitation, perhaps even more severe than the inability to leverage what has been learned in one task to render learning more efficient in a second task, is the inability observed in many RL agents to retain what has been learned when learning a new task---a problem known in machine learning as {\em catastrophic forgetting} \cite{li2018learning,kirkpatrick2017overcoming,kemker2017measuring}. 

In this paper, we take a deeper look at the problems of transfer learning and catastrophic forgetting in the context of deep RL. In particular, we investigate the following research questions: 
\begin{itemize}
\item Compared to single-task learning, can multi-task learning improve an agent's ability to learn a new, similar task? In other words, is an agent trained in multiple tasks more able to transfer learned knowledge from previous tasks to improve the learning of the new one?
\item What is the impact of catastrophic forgetting in multi-task learning? Do existing strategies to address catastrophic forgetting---for example, the Elastic Weight Consolidation algorithm (EWC) \cite{kirkpatrick2017overcoming}---allow a deep RL agent to maintain the acquired knowledge from multiple, previous tasks, after learning new ones?
\end{itemize}
To address the two questions above, we conduct two comparative studies that compare a ``multi-task'' RL agent against specialized ``single-task'' RL agents, both in terms of learning a new task and in terms of catastrophic forgetting. Our results suggest that, in general, multi-task learning provides some advantage when learning a novel (related task); moreover, although the multi-task RL agent suffers from the phenomenon of catastrophic forgetting just like the specialized ``single-task'' agents, our results suggest that the inclusion of a simple mechanism such as EWC can greatly alleviate the impact of such phenomenon.

In our studies we use the {\em asynchronous advantage actor-critic on GPU} (GA3C) algorithm, proposed by \citet{babaeizadeh2016reinforcement}. Our choice of algorithm rests on two key facts. First, GA3C is a well-established deep RL algorithm that has successfully been applied to several benchmark domains from the literature. Secondly, the algorithm can easily be extended to accommodate multi-task learning through a simple modification to its core architecture (see Section~\ref{sect:experiments}). In our second study, we further extend GA3C to include the {\em elastic weight consolidation} (EWC) mechanism \cite{kirkpatrick2017overcoming}. We perform all our experiments using domains from the Atari2600 platform, available as OpenAI Gym environments \cite{brockman2016openai}.

\section{Related Work}
\label{sect:related-work}

In the context of deep RL, several works have explored the problem of transfer learning. For example, \citet{rusu16iclr} propose {\em policy distillation}, where the $Q$-values learned in the target policy are used as regression targets to train a smaller neural network to achieve expert-level performance. In another work, \citet{parisotto16iclr} propose {\em actor-mimic}, which pursues a similar approach but where the learner network seeks to ``mimic'' the policy one or more previously trained expert networks. Finally, \citet{rusu16arxiv} propose {\em progressive networks}, in which the neural network is progressively augmented---through so-called ``columns''---to handle novel tasks, using lateral connections between such columns. One inconvenience of this approach is the scalability, as each column can be seen as a neural network in its own right.

In this paper, we do not use separate networks for different tasks, but instead train a single deep RL agent to simultaneously and continuously perform multiple tasks without forgetting how to perform previously trained ones, in what is known as \emph{continual learning} \cite{parisi2019continual}.

We follow the approach of \citet{birk2017hybrida3c}, who proposed a modification of the asynchronous advantage actor-critic (A3C) algorithm \citep{mnih2016asynchronous}, by having the different actor-learners interact with different environments. With this approach, they seek to test if---by learning two tasks simultaneously---the resulting agent is able to obtain a better policy than two single-task A3C instances, individually trained on each task. Even though we follow their methodology and use two source tasks, an arbitrary number of source tasks can be used.

In our approach, we instead use GA3C \citep{babaeizadeh2016reinforcement}, a modification of the A3C algorithm. Even though newer asynchronous algorithms have surfaced since the release of the A3C and the GA3C \citep{schulman2017proximal,barth2018distributed,hafner2018learning}, we use the GA3C as the foundation for our experiments, as it can easily be extended to accommodate multi-task learning. The GA3C holds a single global network, removing A3C’s need for synchronization. Additionally, it collects the data from the actor-learners differently, in order to take advantage of GPU computation and speedup training. As in the work of \citet{birk2017hybrida3c}, we take advantage of the actor-learners in our approach, allowing them to interact with different environments, naturally enabling multi-task learning.

Unlike the aforementioned works, we are not interested in whether the agent trained in multiple tasks achieves better performance than agents specialized in individual tasks. Instead, our goal is to {\em investigate whether an agent trained on multiple tasks can learn a new task more efficiently}. Our expectation is that, by simultaneously training the agent in several different (although related) tasks, it will acquire more high-level internal representations that may then be useful in the learning of new tasks.

Regarding catastrophic forgetting, several approaches have been explored. We single out three similar algorithms proposed in the literature---namely, EWC \citep{kirkpatrick2017overcoming}, LWF \citep{li2018learning}, and online-EWC \citep{huszar2017quadratic}. According to \citet{kemker2017measuring}, catastrophic forgetting can be observed in sequential learning, whenever a trained model, upon training in a new task, moves abruptly in the space of parameters, effectively ``forgetting'' the original task. To avoid such abrupt changes, when training the network for the second task, these algorithms add an extra term to the loss function of the network that penalizes deviations from the parameters learned for the first task. Following \citet{rusu16arxiv}, \citet{schwarz2018progress} introduce the ``Progress \& Compress'' framework, where each time a new task is learned, a new ``progressive active column'' is created. Afterwards, using the policy distillation method from \citet{rusu16iclr} combined with the EWC algorithm \cite{kirkpatrick2017overcoming}, the parameters are then ``compressed'' into a single ``knowledge base'', updated every time a new task is learnt, whilst preserving older knowledge. As argued by \citet{kirkpatrick2017overcoming}, the EWC is scalable and effective, both in classification and reinforcement learning tasks. In this work, we also {\em investigate whether the use of an approach such as EWC can mitigate catastrophic forgetting while still leveraging the potential advantages of multi-task learning}.


\section{Background}
\label{sect:rl-overview}

We now briefly review the main concepts of RL before moving on to describe our approach.

In reinforcement learning, an agent interacts with an environment in consecutive time steps. At each time step $t$, the agent observes the state $S_t$ and selects an action $A_t$, receiving a reward $r(S_t,A_t)\in\IR$ and transitioning to state $S_{t+1}$. We denote by $\mathcal{S}$ the set of all states, and by $\mathcal{A}$ the set of all actions. The goal of the agent is to select the actions that maximize its {\em expected return}, where the return at time step $t$ is the quantity
\begin{equation*}
R_t=\sum_{\tau=0}^\infty\gamma^\tau r(S_{t+\tau},A_{t+\tau}),
\end{equation*}
where $\gamma\in[0,1)$ is a discount factor. The expected return at time step $t$, $\EE{R_t}$, is a function of the state $S_t$ and of the particular process by which the actions are selected. A {\em policy} $\pi$ is a state-dependent distribution over actions, where $\pi(a\mid s)$ denotes the probability of selecting action $a$ in state $s\in\mathcal{S}$ according to $\pi$. We write
\begin{align*}
V^\pi(s)&=\EE{R_t\mid S_t=s,A_\tau\sim\pi(S_\tau),\tau\geq t},\\
Q^\pi(s,a)&=\EE{R_t\mid S_t=s,A_t=a,A_\tau\sim\pi(S_\tau),\tau>t}.
\end{align*}
A policy $\pi^*$ is optimal if $V^{\pi^*}(s)\geq V^\pi(s)$ for all $\pi$ and all $s\in\mathcal{S}$.

{\em Actor-critic} methods approximate $\pi^*$ by considering a parameterized family of policies $\set{\pi_\theta}$ and updating the parameter $\theta$, at each step $t$, along the gradient of $\EE{R_t}$. Given a trajectory obtained with the current policy $\pi_\theta$, actor-critic methods perform stochastic gradient ascent, using an update of the form
\begin{equation*}
\theta\leftarrow\theta+\alpha\nabla\log\pi_\theta(S_t)(R_t-V^\pi(S_t)).
\end{equation*}
The component performing the policy updates is called the {\em actor}, and the estimation of $V^\pi$ is usually performed independently by a {\em critic}. The quantity $R_t-V^\pi(S_t)$ is an estimate for the {\em advantage function} $A^\pi(s,a)=Q^\pi(s,a)-V^\pi(s,a)$---for which reason several actor-critic algorithms are referred as {\em advantage actor-critic} algorithms---such as A3C and GA3C.

\section{Material and Methods}
\label{sect:experiments}

\begin{figure}[!tb]
    \centering
    \includegraphics[width=\linewidth]{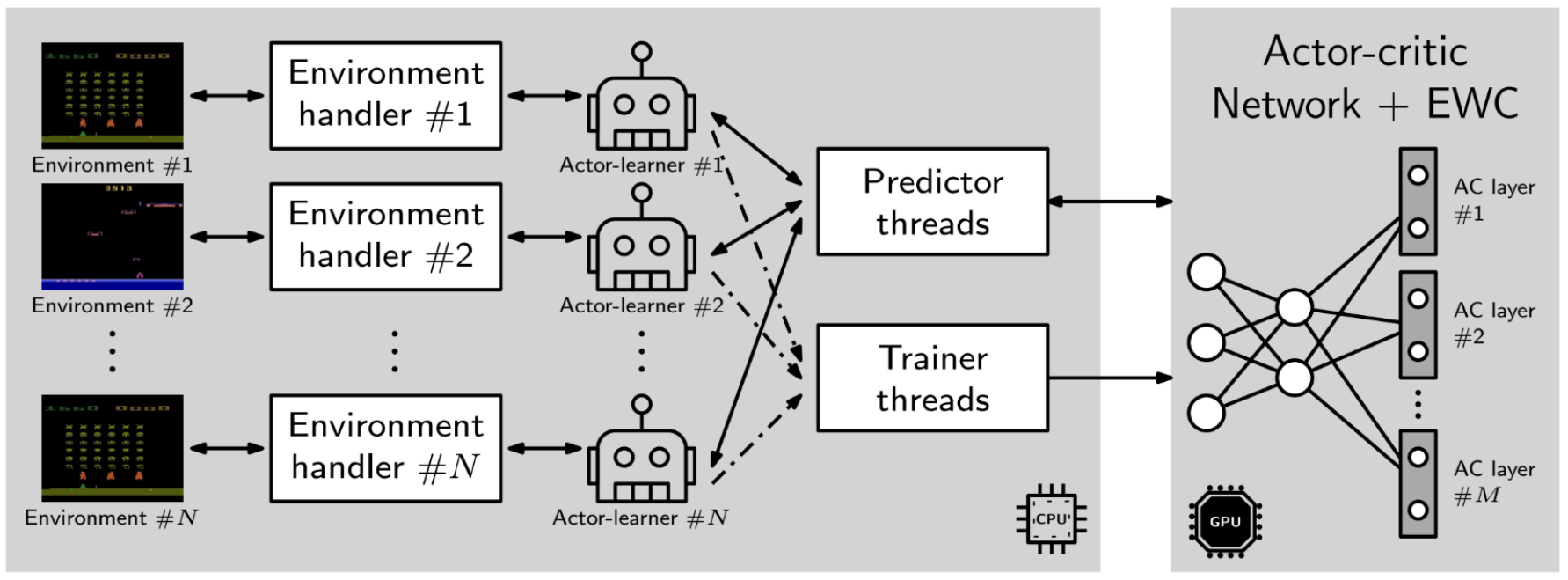}
    \caption{Summarized architecture for the multi-task, EWC-enhanced GA3C. Several actor-learners interact asynchronously with multiple (possibly different) environment instances. As with the original GA3C architecture, each actor-learner submits prediction requests to the actor-critic network using trainer threads. See main text for further network through predictor threads and feed experience batches details.}
    \label{fig:arch}
\end{figure}

In order to test both our hypotheses, we rely on a multi-task version of the GA3C algorithm augmented with EWC. Since the GA3C algorithm has not been designed for either multi-task learning, transfer learning or continual learning, we extend the original GA3C to include:
\begin{itemize}
    \item The hybrid architecture proposed in \cite{birk2017hybrida3c}, in which the actor-learners interact with different environments, enabling multi-task learning;
    \item  The mid-level feature transfer approach from \cite{oquab2014learning}, in which the input layers of a deep convolutional neural network are shared across tasks and only the output layers are task-specific;
    \item The EWC algorithm, by modifying the loss function used to train the actor-critic network with a term that penalizes deviations from parameters learned in previous tasks.
\end{itemize}
Additionally, we also introduce a module called {\em environment handler}, which provides an interface between actor-learners and the environment instances, abstracting the input data into a state $s_{t}$ and a reward signal $r_{t}$. Figure \ref{fig:arch} summarizes the overall architecture.

The algorithm is implemented in Python~3, and the model setup using TensorFlow \cite{abadi2016tensorflow}, running all computational graph forward and backward propagation routines using GPU. 

\subsubsection*{Environment Handlers.}

An \emph{environment handler} is an abstract module responsible for handling the interaction between an actor-learner and an instance of an environment. By abstracting this interaction, we ensure our agent's compatibility with other platforms other than the Atari2600. Even though we open-source only an implementation for the Atari2600 platform, environment handlers enable further studies for multi-task learning with tasks from multiple platforms.

In a given time step $t$, an actor-learner starts by requesting the current state $s_{t}$ and reward $r_{t}$ from its environment handler. The environment handler builds the state $s_{t}$ stacking four observations (frames)---$o_{t}$ , $o_{t-1}$, $o_{t-2}$ and $o_{t-3}$. By stacking four consecutive frames it is possible to capture the temporal context of the current state of the game. This is required since all environments used are history-dependent---i.e. the current observation may have different interpretations, depending on previous observations. The environment handler pre-processes each observation $o_{t}$ (corresponding to a $210\times160\times3$ pixel image for the Atari2600 environment), by resizing it as a grey-scale $84\times84$ image.%
\footnote{We use the same pre-processing method of \citet{babaeizadeh2016reinforcement} which, in turn, was based on the pre-processing method of the DQN \citep{mnih2015human}.}
The resulting $4\times84\times84$ tensor corresponds to the state $s_{t}$ sent to the actor-learner. After obtaining the current state $s_{t}$ from the environment handler, the actor-learner returns an action $a_{t}$, which the environment handler executes upon the environment. After one step, the next state, $s_{t+1}$, and reward, $r_{t}$, are sent to the actor-learner.

\subsubsection*{Loss and back-propagation.}

The loss function used to train the actor-critic network considers the different environments that the agent-learners interact with. We setup a specific loss function, $L_{\mathcal{E}_k}(\theta)$, for each environment $\mathcal{E}_k$. When training using a multi-environment batch, the optimizer uses the corresponding data-points to compute the environment's specific loss, one environment at the time. 

The parameters of the network can be split into 3 sets: $\set{\theta_i}$, $\set{\theta_\pi}$, and $\set{\theta_V}$. The first set, $\set{\theta_i}$, represents the shared parameters for the input network. These parameters are shared by all environments and are updated when back-propagating data-points from all environments. The second set, $\set{\theta_\pi}$, represents the parameters from the actor layers. The parameters associated with each environment are only updated by back-propagating data-points from that specific environment. Finally, the third set, $\set{\theta_V}$, represents the parameters from the critic layers. As with the actor parameters, these are only updated when back-propagating data-points from that specific environment. For each environment, the loss function $L_{\mathcal{E}_k}(\theta)$ combines a loss term associated with the critic and loss term associated with the actor \citep{babaeizadeh2016reinforcement,mnih2016asynchronous}.

Finally, we also introduce EWC to tackle catastrophic forgetting. Suppose that the network has been trained in multiple environments $\mathcal{E}_{1},\ldots,\mathcal{E}_M$ and converged to the set of input parameters $\set{\theta^{*}_i}$. Then, given a new environment, $\mathcal{E}_{M+1}$, we add an additional term to the loss, $L_{\mathcal{E}_{M+1}}(\theta)$, given by
\begin{equation*}
L_\mathrm{EWC}(\theta_i)=\lambda(\theta_i-\theta_i^*)^\top\mathbf{F}(\theta_i-\theta_i^*),  
\end{equation*}
where $\mathbf{F}$ is the Fisher information matrix and can be computed using samples from environments $\mathcal{E}_1,\ldots,\mathcal{E}_M$ \cite{kirkpatrick2017overcoming}. The parameter $\lambda$ allows to specify how important are the old parameters, when learning the new task. We maintain the RMSProp optimizer \cite{ruder2016overview}, using similar parameters to those used by \citet{babaeizadeh2016reinforcement}.

\subsubsection*{Domains}

We resort to the OpenAI Gym toolkit \cite{brockman2016openai}, where available environments from the Atari2600 platform were used as tasks \cite{bellemare2013arcade}. When interacting with an environment, the agent only has access to the game screen (pixels) and current score. We use three different environments consisting of ``bottom-up'' shooting games. Table~\ref{tab:environments} summarizes the used OpenAI Gym environments.%
\footnote{We resort to the deterministic versions of the environments in order to speed up the training.}

\begin{table}[!tb]
\centering
\caption{Environments from OpenAI Gym used in the experiments.}
\label{tab:environments}
\begin{tabular}{c@{\qquad}c}
\toprule
Short name     & OpenAI Gym Environment Name         \\
\midrule
SpaceInvaders  & {\tt SpaceInvadersDeterministic-v4} \\
DemonAttack    & {\tt DemonAttackDeterministic-v4}   \\
Phoenix        & {\tt PhoenixDeterministic-v4}       \\
\bottomrule
\end{tabular}
\end{table}

We then created four agents, using our multi-task GA3C architecture:
\begin{description}
\item[SpaceNet:] GA3C that learned how to play Space Invaders. 24 actor-learners trained on \texttt{SpaceInvaders} for a total of $150,000$ episodes.
\item[DemonNet:] GA3C that learned how to play Demon Attack. 24 actor-learners trained on \texttt{DemonAttack} for a total of $150,000$ episodes.
\item[PhoenixNet:] GA3C that learned how to play Phoenix. 24 actor-learners trained on \texttt{Phoenix} for a total of $150,000$ episodes.
\item[HybridNet:] Multi-task GA3C that learned how to play Space Invaders and Demon Attack. 12 actor-learners asynchronously trained on \text{SpaceInvaders} and \texttt{DemonAttack}, summing up for a total of $150,000$ multi-task episodes.%
\footnote{We do not require each task to have the same number of episodes, since episodes from Space Invaders are quicker than episodes from DemonAttack}.
\end{description}

\section{Results}%

This section describes in detail the results obtained in our two studies.

\subsection{Impact of Multi-Task Learning for New Tasks}

\begin{figure}[!tb]
  \centering
  \subfigure[Space invaders.]{\label{fig:space-150}
    \includegraphics[width=0.45\linewidth]{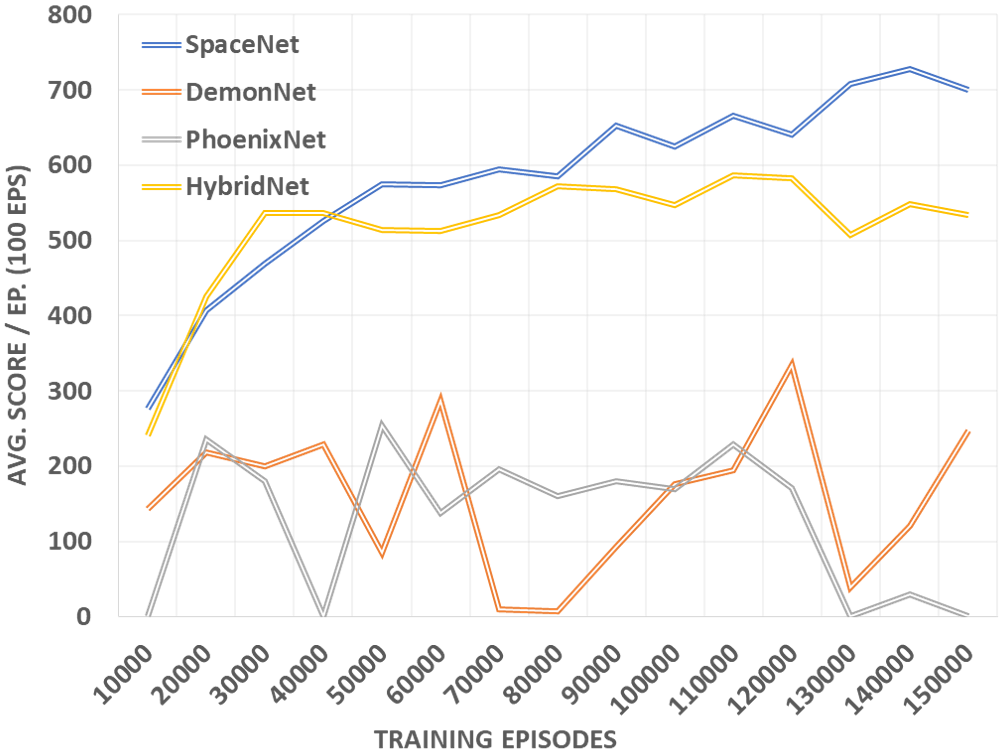}
  } \hfill
  \subfigure[Demon Attack.]{\label{fig:demon-150}
    \includegraphics[width=0.45\linewidth]{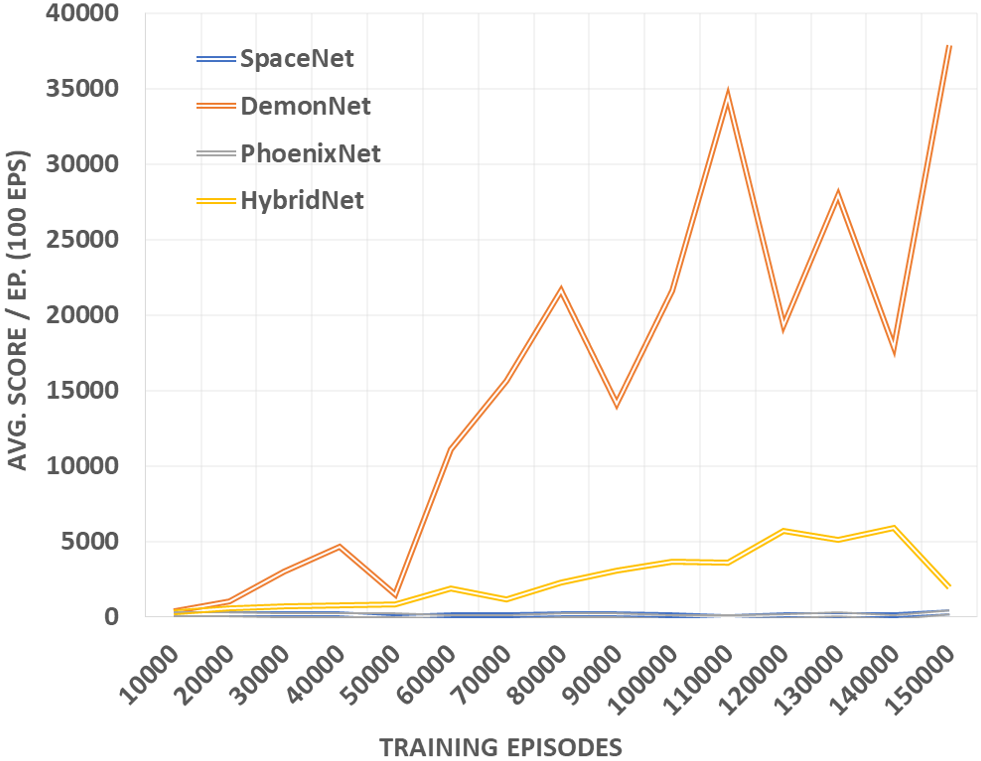}
  }\\
  \subfigure[Phoenix.]{\label{fig:phoenix-150}
    \includegraphics[width=0.45\linewidth]{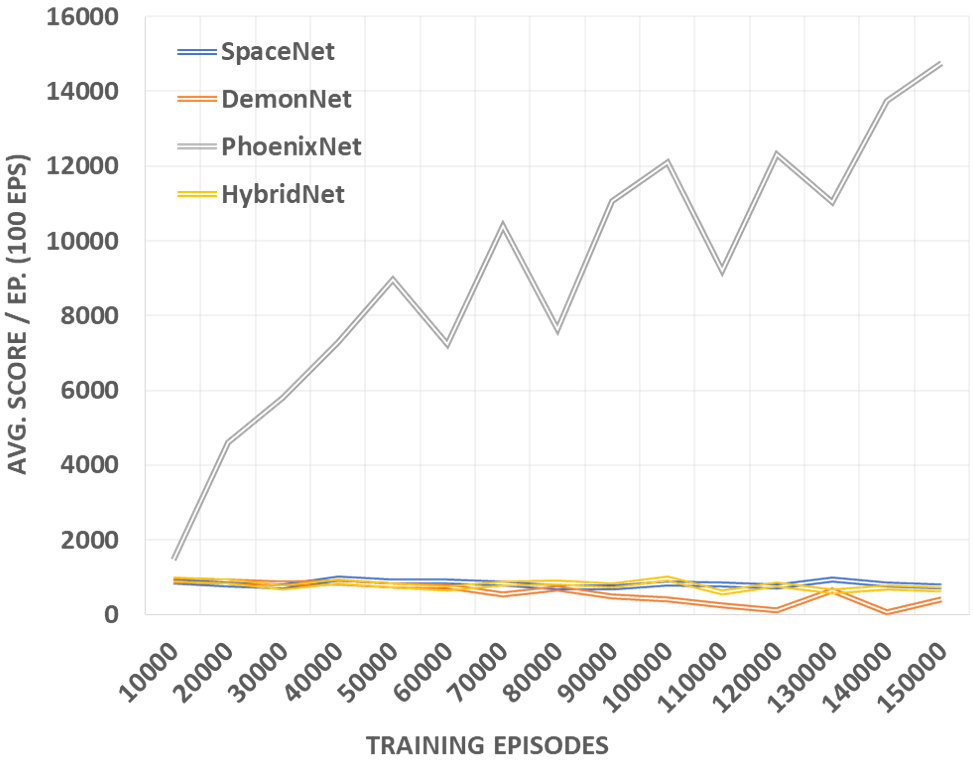}
  }\hfill
  \caption{Learning performance of the four agents in the different environments considered. The results were obtained by evaluating each agent during 100 episodes every $10,000$ training episodes. As expected, the best perform``single-task'' agent attains the best performance.}
  \label{Fig:Single-task-150}
\end{figure}

We start by investigating whether learning multiple tasks simultaneously improves an agent's ability to when learn a new (similar) one. To investigate this hypothesis, we compare the learning performance of HybridNet in a new environment, Phoenix, against the two single-task counterparts (the SpaceNet and the DemonNet). Figure~\ref{Fig:Single-task-150} depicts the comparative performance of all agents in all three environments.

As expected, each ``single-task'' agent outperforms all the others in its corresponding task. The HybridNet, which simultaneously trained for Space Invaders and Demon Attack, was able to obtain a relatively similar performance to the SpaceNet on Space Invaders (Fig.~\ref{fig:space-150}), but was unable to match the performance of the DemonNet on Demon Attack (Fig.~\ref{fig:demon-150}). The difference in performance in Demon Attack may be explained by the discrepancy in the number of training episodes between the two tasks---Demon Attack episodes took longer to finish than episodes from Space Invaders, which means that, in practice, HybridNet played significantly less episodes of Demon Attack than Space Invaders. Finally, the PhoenixNet, which trained for $150,000$ episodes of Phoenix, provides a baseline for the evaluation of the other agents on that environment. We should also reinforce the fact leading to the substantial differences in performance between the HybridNet and the SpaceNet in Space Invaders and the HybridNet and the DemonNet in Demon Attack. When training the HybridNet on both tasks, for a total of 150000 shared episodes, we do not require each task to have the same number of training episodes, since episodes from Space Invaders are quicker than episodes from DemonAttack. Otherwise, the HybridNet would have a higher number of training timesteps for Demon Attack than Space Invaders, resulting in a very task-biased training.

\begin{figure}[!tbp]
  \centering
  \includegraphics[width=0.5\linewidth]{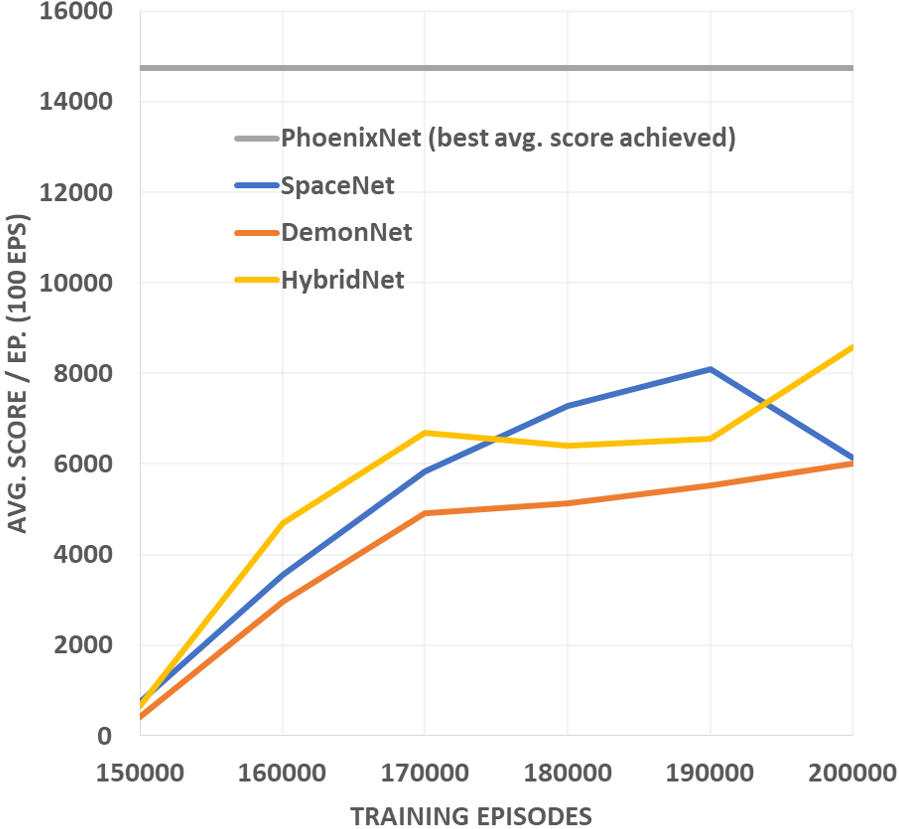}
  \caption{Evaluation of the SpaceNet, DemonNet and HybridNet agents, as they trained on Phoenix for $50,000$ additional episodes. The PhoenixNet performance after training on Phoenix for $150,000$ episodes is provided for reference.}
  \label{fig:phoenix-200}
\end{figure}

We then allowed all three agents (HybridNet, SpaceNet and DemonNet) to learn a new task---Phoenix. We trained all three agents on this new task for an additional $50,000$ episodes. The obtained results are illustrated in Fig.~\ref{fig:phoenix-200}.

The worst performing agent was the DemonNet, with an average score per episode of $6,006.3$, followed by the DemonNet, with a $6,136.6$. At episode 200,000, the HybridNet outperformed the other two, with an average score of $8,587.3$. Even though these results are taken from a single instance of the agent, they suggest that learning from multiple tasks may provide an advantage when learning a new task. It should be also be noted, however, that during learning, the SpaceNet temporarily outperformed HybridNet, at episode 190,000. Since only one instance was trained per agent, this phenomenon may not be statistically significant. However, it is surely an aspect we wish to study further in the future when conducting more extensive research related to the impact of multi-task learning in continuous learning.


\subsection{Catastrophic Forgetting}

\begin{figure}[!tbp]
  \centering
  \subfigure[Space invaders.]{\label{fig:space-150}
    \includegraphics[width=0.45\linewidth]{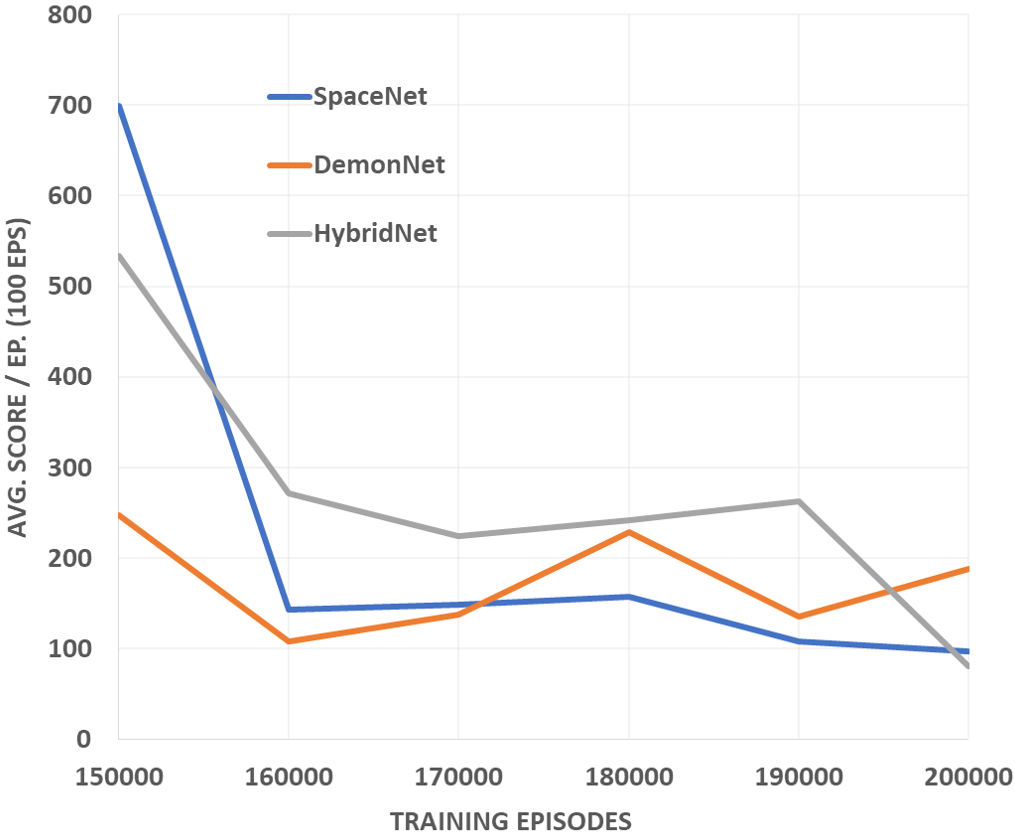}
  } \hfill
  \subfigure[Demon Attack.]{\label{fig:demon-150}
    \includegraphics[width=0.45\linewidth]{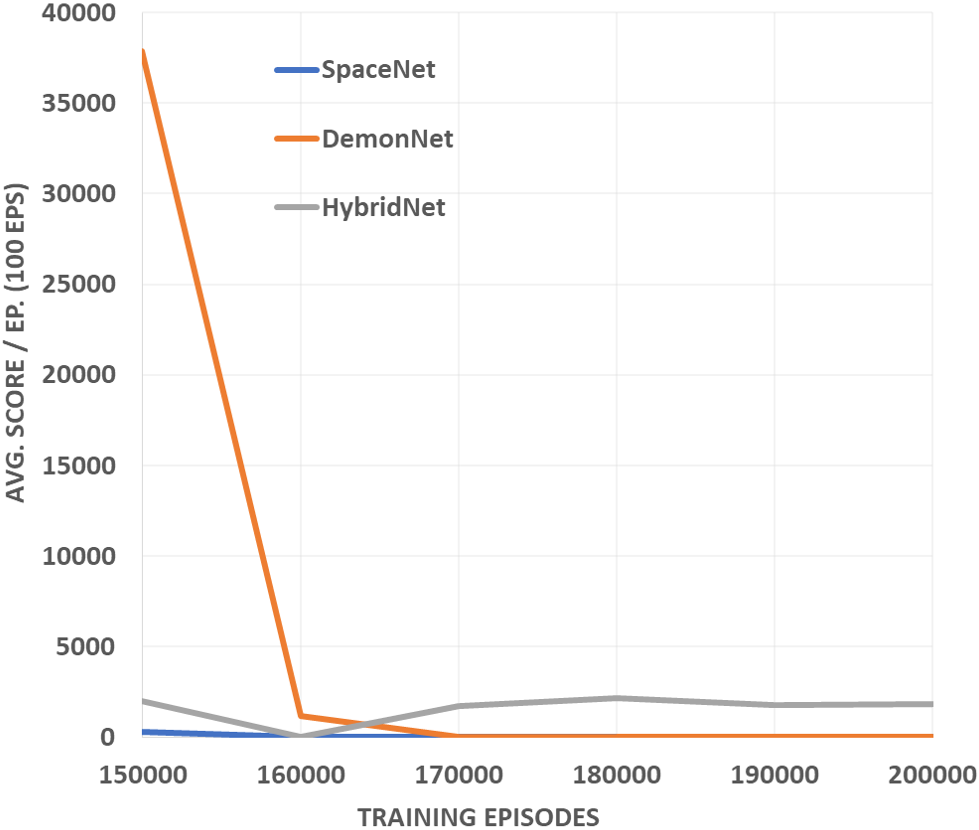}
  }
  \caption{Evaluation of SpaceNet, DemonNet and HybridNet on the two original tasks (Space Invaders and Demon Attack) after training on Phoenix for $50,000$ additional episodes. All agents exhibit catastrophic forgetting, failing to achieve a performance that can be compared with that from Fig.~\ref{Fig:Single-task-150}.}
  \label{Fig:Single-task-200}
\end{figure}

We next considered the problem of catastrophic forgetting. In particular, we took the three agents featured in Fig.~\ref{fig:phoenix-200} and evaluated their the performance in the original tasks. Figure~\ref{Fig:Single-task-200} depicts the performance of all three agents. 

\begin{figure}[!tb]
  \centering
  \subfigure[Space invaders.]{\label{fig:space-150}
    \includegraphics[width=0.45\linewidth]{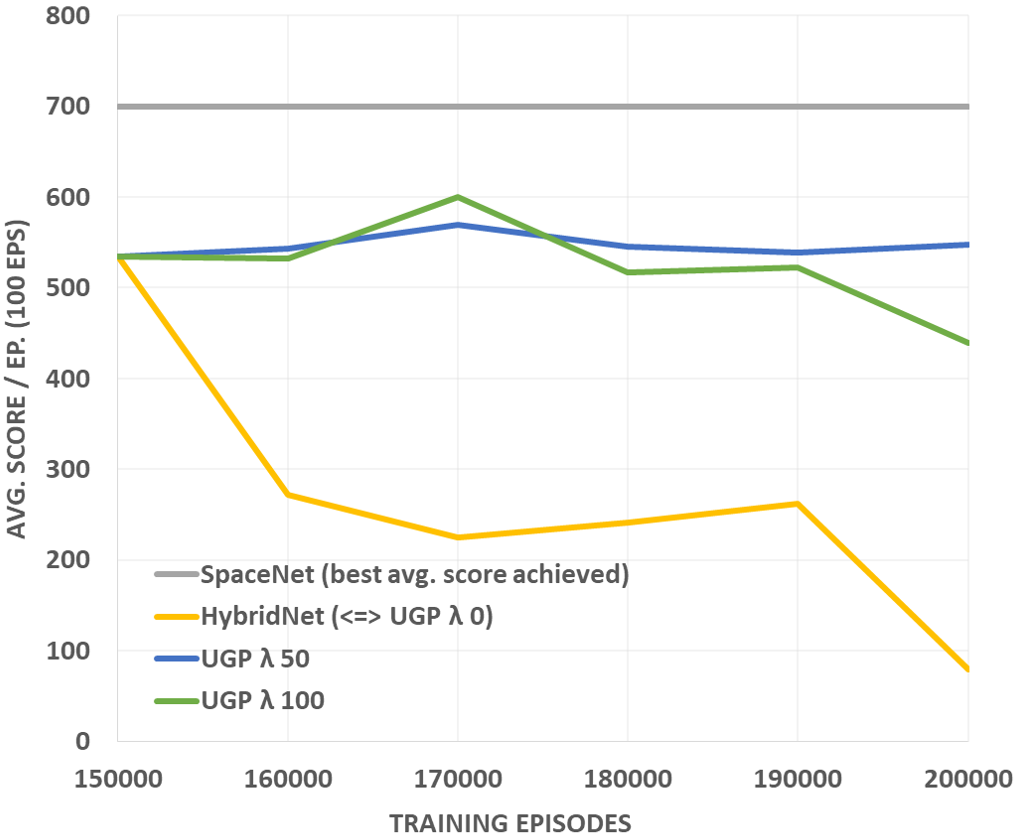}
  } \hfill
  \subfigure[Demon Attack.]{\label{fig:demon-150}
    \includegraphics[width=0.45\linewidth]{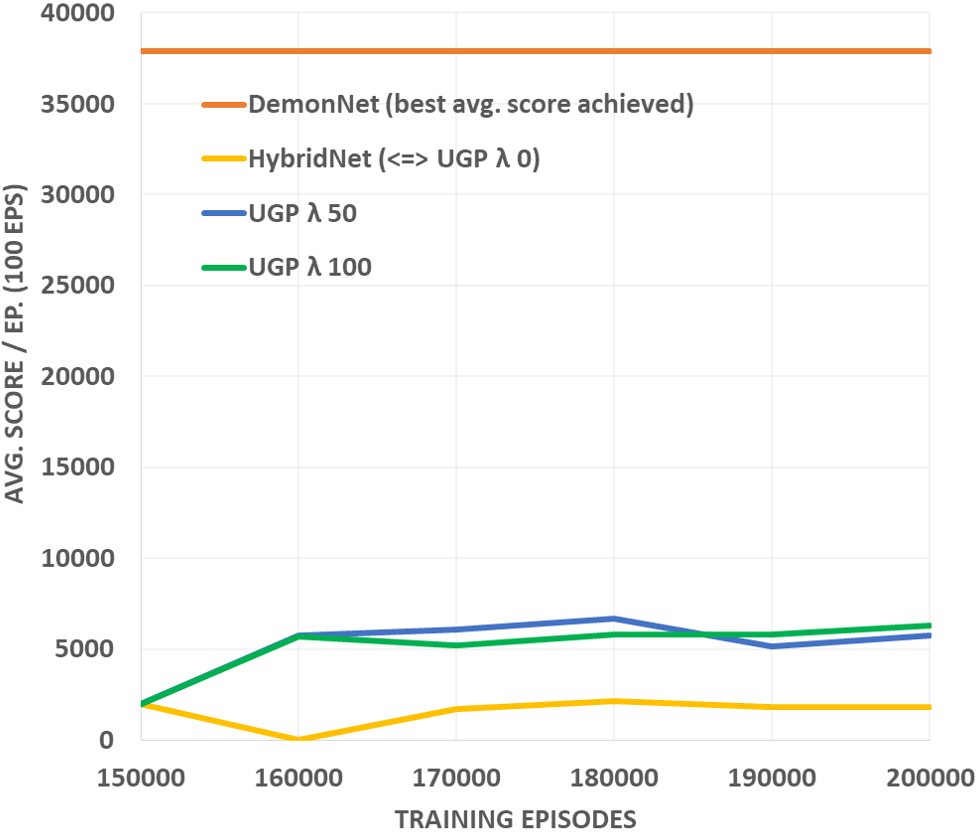}
  }
  \caption{Performance of HybridNet (equivalent to UGP with $\lambda=0.0$), UGP with $\lambda=50.0$ and UGP with $\lambda=100.0$, evaluated on both Space Invaders and Demon Attack, after training $50,000$ additional episodes on Phoenix.}
  \label{Fig:Single-task-200-EWC}
\end{figure}

As expected, after training on Phoenix for $50,000$ episodes, all agents catastrophically forgot how to perform their original tasks. We thus compare the three agents with a version of the ``multi-task'' GA3C HybridNet augmented with EWC. We call the resulting agent the Universal Game Player (UGP), simply for symbolic purposes. We repeated the exact same experiments with three instances of UGP with different values for $\lambda$. In particular, we use $\lambda=0.0$ (in which case UGP is just HybridNet), $\lambda=50.0$ and $\lambda=100.0$. Figure~\ref{Fig:Single-task-200-EWC} depict the performance of UGP.

For $\lambda\in\set{50,100}$ we can observe that, in fact, UGP was able to overcome catastrophic forgetting to a large extent. In the case of the first source task, Space Invaders, all three agents achieved an average score of $533.95$. After $50,000$ additional episodes on Phoenix, the HybridNet achieved an score of $80$ in Space Invaders, while the UGP with $\lambda=50.0$ achieved a score of $547.5$ and the UGP with $\lambda=100$ achieved a score of $439.4$. The differences in performance observed in Demon Attack were also significant, although the final performance of all UGP agents differed more significantly in this second task

\begin{figure}[!tb]
  \centering
  \includegraphics[width=0.5\linewidth]{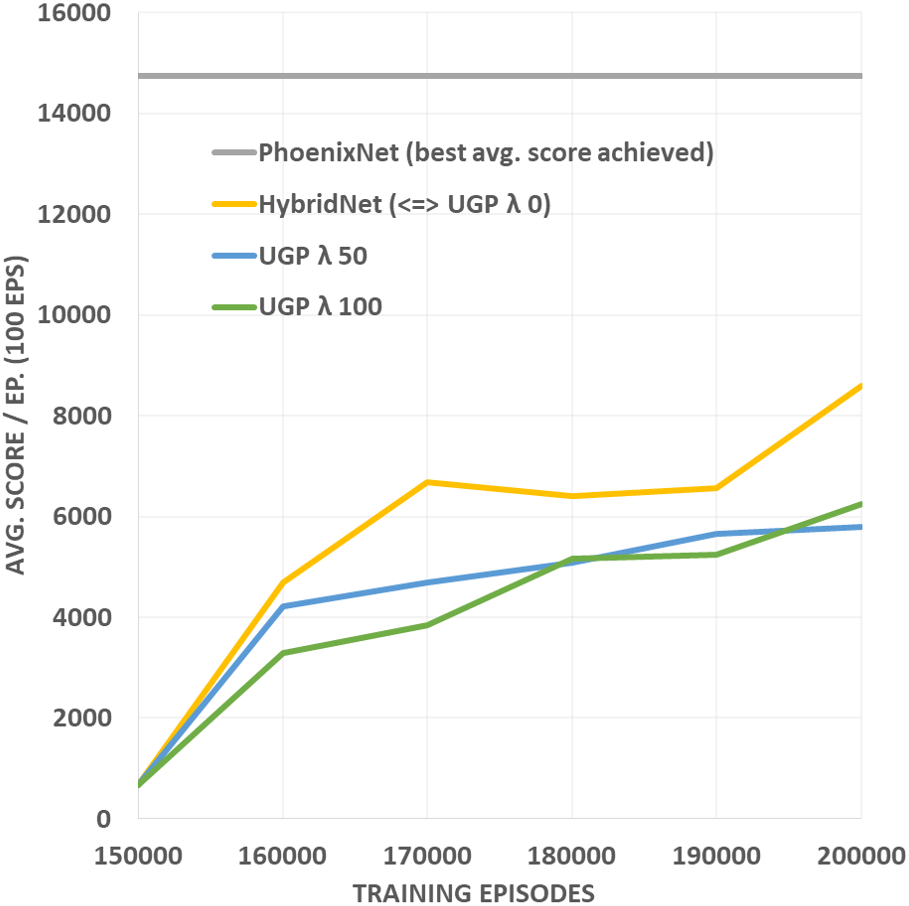}
  \caption{The HybridNet agent (equivalent to an UGP agent without EWC or with EWC $\lambda$=0), UGP agent with $\lambda$=50.0 and UGP agent with $\lambda$=100.0, evaluated on Phoenix, after training 50,000 additional episodes on the environment.}
  \label{fig:phoenix-200-ewc}
\end{figure}

It remains only to assess whether the inclusion of EWC impacted UGP's ability to learn a new task. Figure~\ref{fig:phoenix-200-ewc} shows the result of training UGP in Phoenix.

After training during $50,000$ episodes on Phoenix, the HybridNet was able to achieve an average score $8,587.3$. UGP with $\lambda=50$ achieved an average score of $5,791.1$, while UGP with $\lambda=100$ achieved a score of $6243.4$. As expected, one side-effect of EWC algorithm is the decrease in performance in the new task, since the agent is ``less willing'' to move away from the parameters learned in previous tasks. Nevertheless, UGP with $\lambda=100$ was still able to outperform the single-task networks in learning a new task, confirming that, in fact, multi-task learning can indeed help to learn new tasks, even with the EWC mechanism in place.


\section{Conclusion}
\label{sect:conclusion}
In this work we investigated whether (i) multi-task learning can help in learning new, related tasks and (ii) whether this learning comes at a cost of catastrophic forgetting. Our results show that a modified GA3C algorithm, capable of multi-task learning, can outperform other agents when learning a new, similar task. We also show that, by accommodating the EWC mechanism, the multi-task GA3C is effectively able to mitigate the effects of catastrophic forgetting.

In the future, we plan to study a similar hypothesis, by comparing the effects newer algorithms such as the online-EWC \cite{huszar2017quadratic} and the LWF \cite{li2018learning}, in both single and multi-task learning settings, testing whether multi-task learning is useful in alleviating catastrophic forgetting. 

Thanks to our novel contribution---the use of environment handlers in our architecture---we are now able to apply this methodology to multiple platforms (other than the Atari2600), enabling us to plan and conduct more extensive tests regarding continual learning \cite{parisi2019continual}. The use of environment handlers in our architecture renders such extension straightforward, as it only requires that the platform provides multi-dimensional array of pixels, which can be resized and pre-processed as desired, and a reinforcement measure, which the OpenAI Gym always provides.


\subsection*{Acknowledgments}
\label{sect:acks}

This work was partially supported by national funds through Fundação para a Ciência e a Tecnologia (FCT) with reference UID/CEC/50021/2019, and this material is based upon work supported by the Air Force Office of Scientific Research under award number FA9550-19-1-0020.


\label{sect:bib}
\bibliographystyle{plainnat}
\bibliography{easychair}

\appendix

\end{document}